\documentclass[twocolumn]{bmcart}% uncomment this for twocolumn layout and comment line below
\usepackage{amsthm,amsfonts,amsmath,amssymb}
\RequirePackage{natbib}
\usepackage{nicefrac}    
\usepackage{microtype}
\usepackage{graphicx}
\usepackage{booktabs} 

\RequirePackage[numbers,sort&compress]{natbib}% uncomment this for author-year bibliography
\RequirePackage{url}
\usepackage[utf8]{inputenc} %unicode support
\usepackage[T1]{fontenc}  
\startlocaldefs
\endlocaldefs

\begin{document}

%%% Start of article front matter
\begin{frontmatter}

\begin{fmbox}
\dochead{Research}


\title{Generating Focussed Molecule Libraries for Drug Discovery with Recurrent Neural Networks}

%%%%%%%%%%%%%%%%%%%%%%%%%%%%%%%%%%%%%%%%%%%%%%
%%                                          %%
%% Enter the authors here                   %%
%%                                          %%
%% Specify information, if available,       %%
%% in the form:                             %%
%%   <key>={<id1>,<id2>}                    %%
%%   <key>=                                 %%
%% Comment or delete the keys which are     %%
%% not used. Repeat \author command as much %%
%% as required.                             %%
%%                                          %%
%%%%%%%%%%%%%%%%%%%%%%%%%%%%%%%%%%%%%%%%%%%%%%

\author[
   addressref={aff1},                   % id's of addresses, e.g. {aff1,aff2}
   corref={aff1},                       % id of corresponding address, if any
  % noteref={n1},                        % id's of article notes, if any
   email={marwin.segler@uni-muenster.de}   % email address
]{\inits{MHS}\fnm{Marwin HS} \snm{Segler}}
\author[
   addressref={aff2}
]{\inits{TK}\fnm{Thierry} \snm{Kogej}}
\author[
   addressref={aff3}
]{\inits{CT}\fnm{Christian} \snm{Tyrchan}}
\author[
   addressref={aff4},
   email={waller@shu.edu.cn}
]{\inits{MPW}\fnm{Mark P} \snm{Waller}}

%%%%%%%%%%%%%%%%%%%%%%%%%%%%%%%%%%%%%%%%%%%%%%
%%                                          %%
%% Enter the authors' addresses here        %%
%%                                          %%
%% Repeat \address commands as much as      %%
%% required.                                %%
%%                                          %%
%%%%%%%%%%%%%%%%%%%%%%%%%%%%%%%%%%%%%%%%%%%%%%

\address[id=aff1]{%                           % unique id
  \orgname{Institute of Organic Chemistry \& 
  Center for Multiscale Theory and Computation, Westf\"alische Wilhelms-Universit\"at}, % university, etc
 % \street{Waterloo Road},                     %
  %\postcode{}                                % post or zip code
  \city{M\"unster},                              % city
  \cny{Germany}                                    % country
}
\address[id=aff2]{%
  \orgname{External Sciences, Discovery Sciences, AstraZeneca R\&D Gothenburg},
  \cny{Sweden}
  }
  
\address[id=aff3]{%
  \orgname{Department of Medicinal Chemistry, IMED RIA, AstraZeneca R\&D Gothenburg},
  \cny{Sweden}
}
\address[id=aff4]{%
  \orgname{Department of Physics \& International Centre for Quantum and Molecular Structures,Shanghai University},
  \cny{China}
}

%%%%%%%%%%%%%%%%%%%%%%%%%%%%%%%%%%%%%%%%%%%%%%
%%                                          %%
%% Enter short notes here                   %%
%%                                          %%
%% Short notes will be after addresses      %%
%% on first page.                           %%
%%                                          %%
%%%%%%%%%%%%%%%%%%%%%%%%%%%%%%%%%%%%%%%%%%%%%%

\begin{artnotes}
%\note{Sample of title note}     % note to the article
%\note[id=n1]{...} % note, connected to author
\end{artnotes}

%\end{fmbox}% comment this for two column layout

%%%%%%%%%%%%%%%%%%%%%%%%%%%%%%%%%%%%%%%%%%%%%%
%%                                          %%
%% The Abstract begins here                 %%
%%                                          %%
%% Please refer to the Instructions for     %%
%% authors on http://www.biomedcentral.com  %%
%% and include the section headings         %%
%% accordingly for your article type.       %%
%%                                          %%
%%%%%%%%%%%%%%%%%%%%%%%%%%%%%%%%%%%%%%%%%%%%%%

\begin{abstractbox}

\begin{abstract} % abstract
In \textit{de novo} drug design, computational strategies are used to generate novel molecules with good affinity to the desired biological target. In this work, we show that recurrent neural networks can be trained as generative models for molecular structures, similar to statistical language models in natural language processing. We demonstrate that the properties of the generated molecules correlate very well with the properties of the molecules used to train the model. In order to enrich libraries with molecules active towards a given biological target, we propose to fine-tune the model with small sets of molecules, which are known to be active against that target.

Against Staphylococcus aureus, the model reproduced 14\% of 6051 hold-out test molecules that medicinal chemists designed, whereas against Plasmodium falciparum (Malaria) it reproduced 28\% of 1240 test molecules. When coupled with a scoring function, our model can perform the complete \textit{de novo} drug design cycle to generate large sets of novel molecules for drug discovery.
\end{abstract}

%%%%%%%%%%%%%%%%%%%%%%%%%%%%%%%%%%%%%%%%%%%%%%
%%                                          %%
%% The keywords begin here                  %%
%%                                          %%
%% Put each keyword in separate \kwd{}.     %%
%%                                          %%
%%%%%%%%%%%%%%%%%%%%%%%%%%%%%%%%%%%%%%%%%%%%%%


% MSC classifications codes, if any
%\begin{keyword}[class=AMS]
%\kwd[Primary ]{}
%\kwd{}
%\kwd[; secondary ]{}
%\end{keyword}

\end{abstractbox}
\end{fmbox}% uncomment this for twcolumn layout

\end{frontmatter}

%%%%%%%%%%%%%%%%%%%%%%%%%%%%%%%%%%%%%%%%%%%%%%
%%                                          %%
%% The Main Body begins here                %%
%%                                          %%
%% Please refer to the instructions for     %%
%% authors on:                              %%
%% http://www.biomedcentral.com/info/authors%%
%% and include the section headings         %%
%% accordingly for your article type.       %%
%%                                          %%
%% See the Results and Discussion section   %%
%% for details on how to create sub-sections%%
%%                                          %%
%% use \cite{...} to cite references        %%
%%  \cite{koon} and                         %%
%%  \cite{oreg,khar,zvai,xjon,schn,pond}    %%
%%  \nocite{smith,marg,hunn,advi,koha,mouse}%%
%%                                          %%
%%%%%%%%%%%%%%%%%%%%%%%%%%%%%%%%%%%%%%%%%%%%%%

%%%%%%%%%%%%%%%%%%%%%%%%% start of article main body
% <put your article body there>

%%%%%%%%%%%%%%%%
%% Background %%
%%

\section{Introduction}
Chemistry is the language of nature. Chemists speak it fluently and have made their discipline one of the true contributors to human well-being, which has \textit{``change[d] the way you live and die''}.\cite{whitesides2015reinventing} 
This is particularly true for medicinal chemistry. However, creating novel drugs is an extraordinarily hard and complex problem.\cite{schneider2016novo} 
One of the many challenges in drug design is the sheer size of the search space for novel molecules. It has been estimated that $10^{60}$ drug-like molecules could possibly be synthetically accessible.\cite{reymond2012enumeration} 
Chemists have to select and examine molecules from this large space to find molecules that are active towards a biological target. Active means for example that a molecule binds to a biomolecule, which causes an effect in the living organism, or inhibits replication of bacteria.
Modern high-throughput screening techniques allow to test molecules in the order of 10$^6$ in the lab.\cite{schneider2008molecular} However, larger experiments will get prohibitively expensive. Given this practical limitation of \textit{in vitro} experiments, it is desirable to have computational tools to narrow down the enormous search space. 
\textit{Virtual screening} is a commonly used strategy to search for promising molecules amongst millions of existing or billions of virtual molecules.\cite{stumpfe2011similarity}  
Searching can be carried out using similarity-based metrics, which provides a quantifiable numerical indicator of closeness between molecules. 
In contrast, in \textit{de-novo} drug design, one aims to directly create novel molecules that are active towards the desired biological target.\cite{schneider2005computer,hartenfeller2011enabling} Here, like in any molecular design task, the computer has to
\begin{enumerate}
\item[i] create molecules,
\item[ii] score and filter them, and
\item[iii] search for better molecules, building on the knowledge gained in the previous steps.
\end{enumerate}

Task i, the generation of novel molecules, is usually solved with one of two different protocols.\cite{hartenfeller2011enabling} One strategy is to build molecules from predefined groups of atoms or fragments. Unfortunately, these approaches often lead to molecules that are very hard to synthesise.\cite{hartenfeller2012dogs} Therefore, another established approach is to conduct virtual chemical reactions based on expert coded rules, with the hope that these reactions could then also be applied in practice to make the molecules in the laboratory.\cite{hartenfeller2011collection}
These systems give reasonable drug-like molecules, and are considered as ``the solution'' to the structure generation problem.\cite{schneider2016novo} We generally share this view. However, we have recently shown that the predicted reactions from these rule-based expert systems can sometimes fail.\cite{neural-symbolic} Also, focussing on a small set of robust reactions can unnecessarily restrict the possibly accessible chemical space.

Task ii, scoring molecules and filtering out undesired structures, can be solved with substructure filters for undesirable reactive groups in conjunction with established approaches such as docking\cite{kitchen2004docking} or machine-learning (ML) approaches.\cite{hartenfeller2011enabling,varnek2012machine,mitchell2014machine} The ML approaches are split into two branches: Target prediction classifies molecules into active and inactive, and quantitative structure-activity relationships (QSAR) seek to quantitatively predict a real-valued measure for the effectiveness of a substance (as a regression problem). As molecular descriptors, Signature Fingerprints, Extended-Connectivity (ECFP) and atom pair (APFP) fingerprints and their fuzzy variants are the \textit{de-facto} standard today.\cite{riniker2013open,rogers2010extended,alvarsson2014ligand} Convolutional Networks on Graphs are a recent addition to the field of molecular descriptors.\cite{duvenaud2015convolutional,kearnes2016molecular}
Random Forests and Neural Networks are currently the most widely used machine learning models for target prediction.\cite{zupan1991neural,gasteiger1993neural,zupan1999neural,lusci2013deep,unterthiner2014deep,unterthiner2015toxicity,schneider2016hybrid,gawehn2016deep,ramsundar2015massively,kearnes2016modeling,behler2015constructing,behler2007generalized,ma2015deep}

This leads to task iii, the search for molecules with the right binding affinity combined with optimal molecular properties. In earlier work, this was performed (among others) with classical global optimisation techniques, for example genetic algorithms or ant-colony optimisation.\cite{hartenfeller2011enabling,reutlinger2014multi} Furthermore, \textit{de novo} design is related to inverse QSAR.\cite{miyao2010exhaustive,miyao2016inverse,takeda2016chemical,mishima2014development} While in \textit{de novo} design design, a \textit{regular} QSAR mapping $X \rightarrow y$ from molecular descriptor space $X$ to properties $y$ is used as the scoring function for the global optimizer, in \textit{inverse} QSAR one aims to find an explicit inverse mapping $y \rightarrow X$, and then maps back from optimal points in descriptor space $X$ to valid molecules. However, this is not well defined, because molecules are inherently discrete. Several protocols have been developed to address this, for example enumerating all structures within the constraints of hyper-rectangles in the descriptor space.\cite{white2010generative,miyao2010exhaustive,miyao2016inverse,takeda2016chemical,mishima2014development,patel2009knowledge} G\'omez-Bombarelli \textit{et al.} proposed to learn continuous representations of molecules with variational auto-encoders, based on the model by Bowman \textit{et al.},\cite{bowman2015generating} and to perform Bayesian optimisation in this vector space to optimise molecular properties.\cite{gomez2016automatic} Nevertheless, this approach was not applied to create active drug molecules, and did not succeed in optimising more complex molecular properties, such as emission color and delayed fluorescence decay rate ($k_\text{TADF}$).\cite{gomez2016automatic}

In this work, we suggest a novel, completely data-driven \textit{de novo} drug design approach. It relies only on a generative model for molecular structures, based on a recurrent neural network, that is trained on large sets of molecules. Generative models learn a probability distribution over the training examples; sampling from this distribution generates new examples similar to the training data. Intuitively, a generative model for molecules trained on drug molecules would "know" how valid and reasonable drug-like molecules look like, and could be used to generate more drug-like molecules. However, for molecules, these models have been studied rarely, and rigorously only with traditional models such as Gaussian mixture models (GMM).\cite{voss,white2010generative} 
Recently, recurrent neural networks (RNNs) have emerged as powerful generative models in very different domains, such as natural language processing,\cite{jozefowicz2016exploring}  speech,\cite{graves2004biologically} images,\cite{van2016pixel} video,\cite{srivastava2015unsupervised} formal languages,\cite{gers2001lstm} computer code generation,\cite{bhoopchand2016learning} and music scores.\cite{eck2002finding} In this work, we highlight the analogy of language and chemistry, and show that RNNs can also generate reasonable molecules. Furthermore, we demonstrate that RNNs can also transfer their learned knowledge from large molecule sets to directly produce novel molecules that are biologically active by retraining the models on small sets of already known actives. We test our models by reproducing hold-out test sets of known biologically active molecules.

\section{Methods}
\subsection{Representing Molecules}

\begin{figure}[htbp]
\begin{center}
\includegraphics[width=0.9\linewidth]{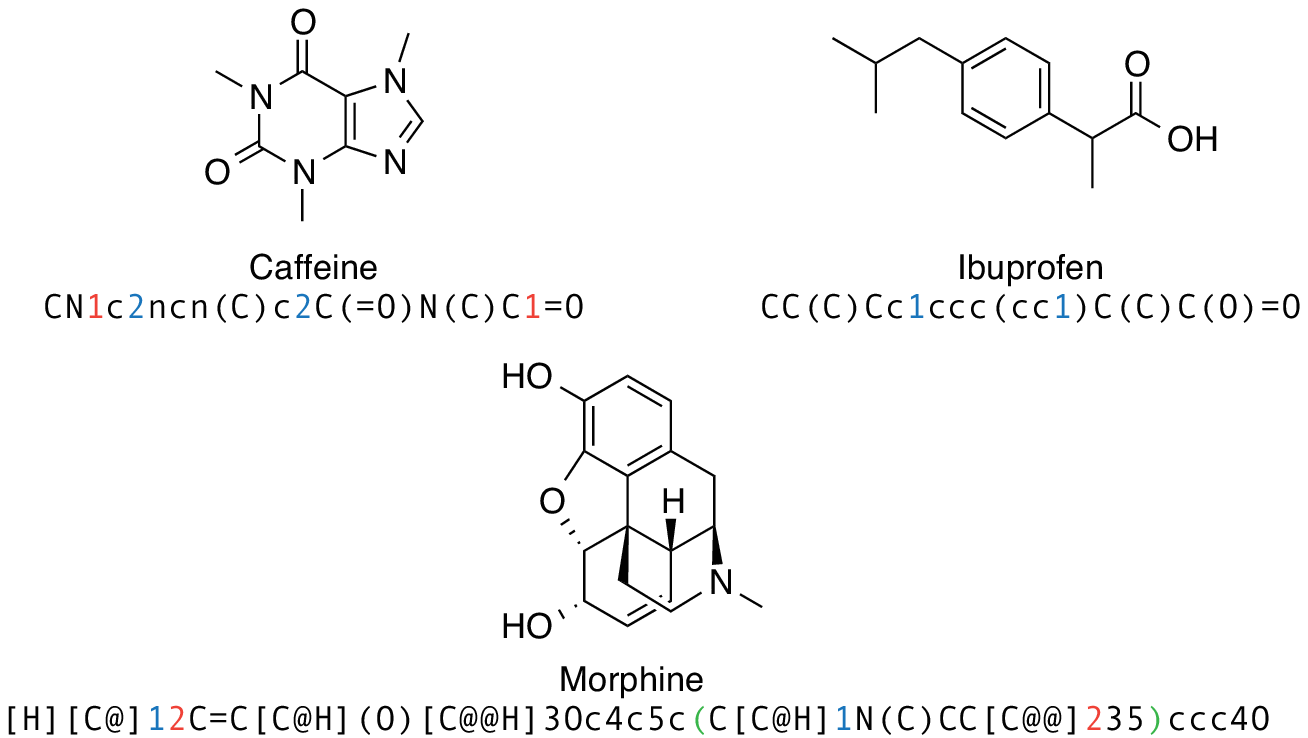}
\caption{Examples of molecules and their \textsc{Smiles} representation. To correctly create smiles, the model has to learn long term dependencies, for example to close rings (indicated by numbers) and brackets.}
\label{fig:smiles}
\end{center}
\end{figure}

To connect chemistry with language, it is important to understand how molecules are represented. Usually, they are modeled by molecular graphs, also called Lewis structures in chemistry. In molecular graphs, atoms are labeled nodes. The edges are the bonds between atoms, which are labeled with the bond order (e.g. single, double or triple). 
One could therefore envision having a model that reads and outputs graphs. Several common chemistry formats store molecules in such a manner. However, in models for natural language processing, the input and output of the model are usually sequences of single letters, strings or words. 
We therefore employ the \textsc{Smiles} format, which encodes molecular graphs compactly as human-readable strings. \textsc{Smiles} is a formal grammar which describes molecules with an alphabet of characters, for example \texttt{c} and \texttt{C} for aromatic and aliphatic carbon atoms, \texttt{O} for oxygen, \texttt{-}, \texttt{=} and \texttt{\#} for single, double and triple bonds (see Figure \ref{fig:smiles}).\cite{weininger1988smiles} 
To indicate rings, a number is introduced at the two atoms where the ring is closed. For example, benzene in aromatic \textsc{Smiles} notation would be \texttt{c1ccccc1}. Side chains are denoted by round brackets. 
To generate valid \textsc{Smiles}, the generative model would have to learn the \textsc{Smiles} grammar, which includes keeping track of rings and brackets to eventually close them. In morphine, a complex natural product, the number of steps between the first \texttt{1} and the second \texttt{1}, indicating a ring, is 32. Having established a link between molecules and (formal) language, we can now discuss language models.

\subsection{Language Models and Recurrent Neural Networks}

Given a sequence of words $(w_1,...,w_i)$, language models predict the distribution of the $(i+1)$th word $w_{i+1}$.\cite{yoav} For example, if a language model receives the sequence \texttt{"Chemistry is"}, it would assign different probabilities to possible next words. \texttt{"fascinating"}, \texttt{"important"}, or \texttt{"challenging"} would receive high probabilities, while \texttt{"runs"} or \texttt{"potato"} would receive very low probabilities. Language models can both capture the grammatical correctness ("runs" in this sentence is wrong) and the meaning ("potato" does not make sense). Language models are implemented for example in message autocorrection in many modern smartphones. Interestingly, language models do not have to use words. They can also be based on characters or letters.\cite{yoav} In that case, when receiving the sequence of characters \texttt{chemistr}, it would assign a high probability to \texttt{y}, but a low probability to \texttt{q}.
To model molecules instead of language, we simply swap words or letters with atoms, or, more practically, characters in the \textsc{Smiles} alphabet, which form a (formal) language. For example, if the model receives the sequence \texttt{c1ccccc}, there is a high probability that the next symbol would be a \texttt{"1"}, which closes the ring, and yields benzene.

More formally, to a sequence $S$ of symbols $s_i$ at steps $t_i \in T$, the language model assigns a probability of
\begin{equation}
P_\theta(S) = P_\theta(s_1) \cdot \prod^T_{t=2}P_\theta(s_t|s_{t-1},...,s_1) \label{eq:prod}
\end{equation}
where the parameters $\theta$ are learned from the training set.\cite{yoav}
In this work, we use a recurrent neural network (RNN) to estimate the probabilities of Equation \ref{eq:prod}.
In contrast to regular feedforward neural networks, RNNs maintain state, which is needed to keep track of the symbols seen earlier in the sequence. In abstract terms, an RNN takes a sequence of input vectors $\mathbf{x}_{1:n} = (\mathbf{x}_1,...,\mathbf{x}_n)$ and an initial state vector $\mathbf{h}_0$, and returns a sequence of state vectors $\mathbf{h}_{1:n} = (\mathbf{h}_1,...,\mathbf{h}_n)$ and a sequence of output vectors $\mathbf{y}_{1:n} = (\mathbf{y}_1,...,\mathbf{y}_n)$.
The RNN consists of a recursively defined function $R$, which takes a state vector $\mathbf{h}_{i}$ and input vector  $\mathbf{x}_{i+1}$ and returns a new state vector $\mathbf{h}_{i+1}$. Another function $O$ maps a state vector $\mathbf{h}_{i}$ to an output vector $\mathbf{y}_{i}$.\cite{yoav}

\begin{align}
\text{RNN}(\mathbf{h}_0, \mathbf{x}_{1:n}) &= \mathbf{h}_{1:n}, \mathbf{y}_{1:n}\\
\mathbf{h}_i &= R(\mathbf{h}_{i-1},\mathbf{x}_i)\\
\mathbf{y}_i &= O(\mathbf{h}_{i})
\end{align}

The state vector $\mathbf{h}_{i}$ stores a representation of the information about all symbols seen in the sequence so far.
\begin{figure}[htbp]
\begin{center}
\includegraphics[width=0.9\linewidth]{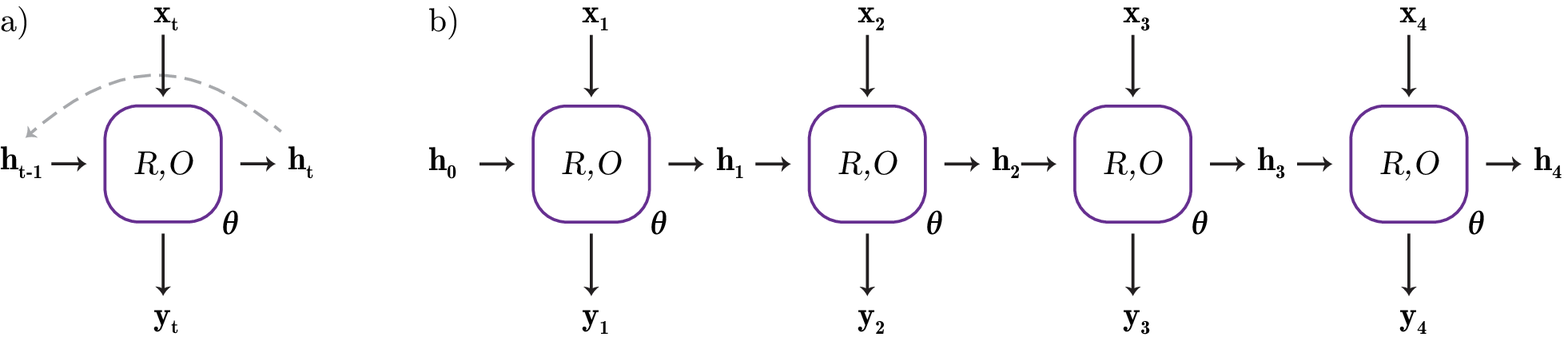}
\caption{a) Recursively defined RNN b) The same RNN, unrolled. The parameters $\theta$ (the weight matrices of the neural network) are shared over all time steps.}
\label{fig:rnn}
\end{center}
\end{figure}
As an alternative to the recursive definition, the recurrent network can also be \textit{unrolled} for finite sequences (see Figure \ref{fig:rnn}). An unrolled RNN can be seen as a very deep neural network, in which the parameters $\theta$ are shared among the layers, and the hidden state $\mathbf{h}_t$ is passed as an additional input to the next layer. Training the unrolled RNN to fit the parameters $\theta$ can then simply be done by using backpropagation to compute the gradients with respect to the loss function, which is categorical cross-entropy in this work.\cite{yoav}

As the specific RNN function, in this work, we use the Long Short Term Memory (LSTM), which was introduced by Hochreiter and Schmidhuber.\cite{hochreiter1997long} It has been used successfully in many natural language processing tasks,\cite{jozefowicz2016exploring} for example in Google's Neural Machine Translation system.\cite{johnson2016google}  
For excellent in-depth discussions of the LSTM, we refer to the articles by Goldberg,\cite{yoav} Graves,\cite{graves2013generating} Olah,\cite{colah} and Greff \textit{et al}.\cite{greff2015lstm}

To encode the \textsc{Smiles} symbols as input vectors $\mathbf{x}_t$ , we employ the "one-hot" representation.\cite{graves2013generating} This means if there are $K$ symbols, and $k$ is the symbol to be input at step $t$, then we can construct an input vector $\mathbf{x}_t$ with length $K$, whose entries are all zero except the $k$-th entry, which is one. If we assume a very restricted set of symbols \{\texttt{c}, \texttt{1}, \texttt{\textbackslash n}\}, input \texttt{c} would correspond to $\mathbf{x}_t=(1,0,0)$, \texttt{1} to  $\mathbf{x}_t=(0,1,0)$ and \texttt{\textbackslash n} to $\mathbf{x}_t=(0,0,1)$. 
\begin{figure*}[htb]
\begin{center}
\includegraphics[width=0.7\textwidth]{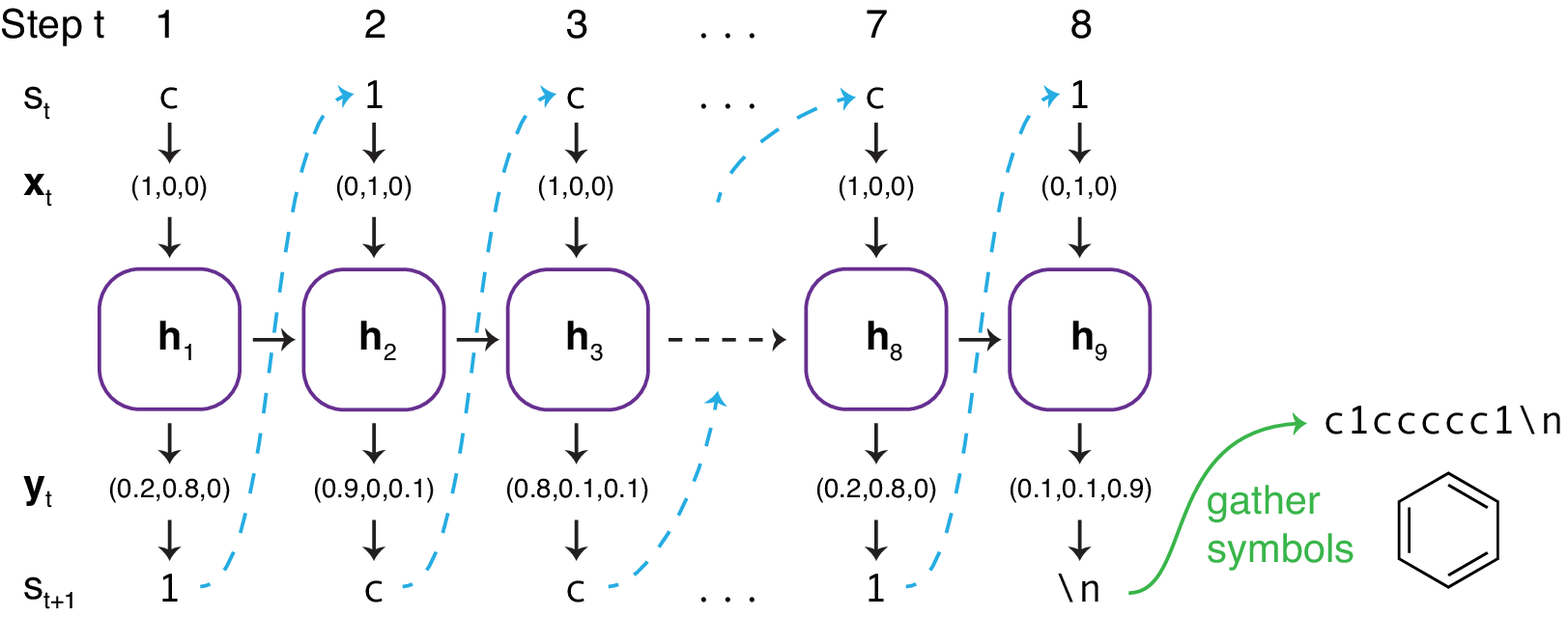}
\caption{The Symbol Generation and Sampling Process. We start with a random seed symbol $\mathbf{s}_1$, here \texttt{c}, which gets converted into a one-hot vector $\mathbf{x}_1$ and input into the model. The model then updates its internal state $\mathbf{h}_0$ to $\mathbf{h}_1$ and outputs $\mathbf{y}_1$, which is the probability distribution over the next symbols. Here, sampling yields $\mathbf{s}_2=$\texttt{1}. Converting $\mathbf{s}_2$ to $\mathbf{x}_2$, and feeding it to the model leads to updated hidden state $\mathbf{h}_2$ and output $\mathbf{y}_2$, from which can sample again. This iterative symbol-by-symbol procedure can be continued as long as desired. In this example, we stop it after observing an EOL (\texttt{\textbackslash n}) symbol, and obtain the \textsc{Smiles} for benzene. The hidden state $\mathbf{h}_i$ allows the model to keep track of opened brackets and rings, to ensure that they will be closed again later.}
\label{fig:gentext}
\end{center}
\end{figure*}

The probability distribution $P_\theta(s_{t+1}| s_{t},...,s_1)$ of the next symbol given the already seen sequence is thus a multinomial distribution, which is estimated using the output vector $\mathbf{y}_t$ of the recurrent neural network at time step $t$ by
\begin{equation}
P_\theta(s_{t+1} = k | s_{t},...,s_1) = \frac{\exp( y^k_t ) }{\sum^K_{k^\prime=1} \exp(y^{k^\prime}_t)}\label{eq:softmax}
\end{equation}
where $y^k_t$ corresponds to the $k$-th element of vector $\mathbf{y}_t$.\cite{graves2013generating} 
Sampling from this distribution would then allow generating novel molecules: After sampling a \textsc{Smiles} symbol $s_{t+1}$ for the next time step $t+1$, we can construct a new input vector $\mathbf{x}_{t+1}$, which is fed into the model, and via $\mathbf{y}_{t+1}$ and Equation \ref{eq:softmax} yields $P_\theta(s_{t+2}| s_{t+1},...,s_1)$. Sampling from the latter generates $s_{t+2}$, which serves again also as the model's input for the next step (see Figure \ref{fig:gentext}). This symbol-by-symbol sampling procedure is repeated until the desired number of characters has been generated.\cite{graves2013generating} 

To indicate that a molecule is "completed", each molecule in our training data finishes with an ``end of line'' (EOL) symbol, in our case the single character \texttt{\textbackslash n} (which means the training data is just a simple \textsc{Smiles} file). Thus, when the system outputs an EOL, a generated molecule is finished. However, we simply continue sampling, thus generating a regular \textsc{Smiles} file that contains one molecule per line.

In this work, we used a network with three stacked LSTM layers, using the keras library.\cite{chollet2015keras} 
The model was trained with back propagation through time,\cite{graves2013generating}  using the ADAM optimizer at standard settings.\cite{kingma2014adam} To mitigate the problem of exploding gradients during training, a gradient norm clipping of 5 is applied.\cite{graves2013generating}

\subsection{Transfer Learning}
For many machine learning tasks, only small datasets are available, which might lead to overfitting with powerful models such as neural networks. In this situation, \textit{transfer learning} can help.\cite{cirecsan2012transfer} Here, a model is first trained on a large dataset for a different task. Then, the model is retrained on the smaller dataset, which is also called \textit{fine-tuning}. The aim of transfer learning is to learn general features on the bigger data set, which also might be useful for the second task in the smaller data regime. To generate focussed molecule libraries, we first train on a large, general set of molecules, then perform fine-tuning on a smaller set of specific molecules, and after that start the sampling procedure.

\subsection{Target Prediction}
To verify whether the generated molecules are active on the desired targets, standard target prediction was employed. 
Machine learning-based target prediction aims to learn a classifier $c:M\rightarrow \{1,0\}$ to decide whether a molecule $m \in$ molecular descriptor space $M$ is active or not against a target.\cite{mitchell2014machine,varnek2012machine}  
The molecules are split into actives and inactives using a threshold on a measure for the substance effectiveness. \textit{p}IC$_{50} = -\log_{10}$(IC$_{50}$) is one of the most widely used metrics for this purpose. IC$_{50}$ is the \textit{half maximal inhibitory concentration}, that is the concentration of drug that is required to inhibit 50\% of a biological target's function \textit{in vitro}.

To predict whether the generated molecules are active towards the biological target of interest, target prediction models (TPMs) were trained for all the tested targets (5-HT$_{\text{2A}}$, \textit{Plasmodium falciparum} and \textit{Staphylococcus aureus}). We evaluated Random Forest, Logistic Regression, (Deep) Neural Networks and Gradient Boosting Trees (GBT) as models with ECFP4 (Extended Connectivity Fingerprint with a diameter of 4) as the molecular descriptor.\cite{riniker2013open,rogers2010extended} We found that GBTs slightly outperformed all other models, and used these as our virtual assay in all studies (see Supporting Information for details). ECFP4 fingerprints were generated with CDK version 1.5.13.\cite{steinbeck2006recent,steinbeck2003chemistry} Scikit-Learn,\cite{scikit-learn} xgBoost\cite{chen2016xgboost} and keras\cite{chollet2015keras} were used as the machine learning libraries. For 5-HT$_{\text{2A}}$ and \textit{Plasmodium}, molecules are considered as active for the TPM if their IC$_{50}$ reported in ChEMBL is < 100 n\textsc{m}, which translates to a \textit{p}IC$_{50}$ > 7, whereas for \textit{Staphylococcus}, we used \textit{p}MIC > 3.

\subsection{Data}
The chemical language model was trained on a \textsc{Smiles} file containing 1.4 million molecules from the ChEMBL database, which contains molecules and measured biological activity data. 
The \textsc{Smiles} strings of the molecules were canonicalized (which means finding a unique representation that is the same for isomorphic molecular graphs)\cite{weininger1989smiles,canon1} before training with the CDK chemoinformatics library, yielding a \textsc{Smiles} file that contained one molecule per line.\cite{steinbeck2006recent,steinbeck2003chemistry} 
It has to be noted that ChEMBL contains many peptides, natural products with complex scaffolds, Michael acceptors, benzoquinones, hydroxylamines, hydrazines etc. which is reflected in the generated structures (see below). 
This corresponds to 72 million individual characters, with a vocabulary size of 51 unique characters. 51 characters is only a subset of all \textsc{Smiles} symbols, since the molecules in ChEMBL do not contain many of the heavy elements. As we have to set the number of symbols as a hyperparameter during model construction, and the model can only learn the distribution over the symbols present in the training data, this implies that only molecules with these 51 \textsc{Smiles} symbols seen during training can be generated during sampling.

The 5-HT$_{\text{2A}}$, the \textit{Plasmodium falciparum} and the \textit{Staphylococcus aureus} dataset were also obtained from ChEMBL. The molecules for the hold-out test sets were removed from the training data.

\subsection{Model Evaluation}
To evaluate the models for a test set $T$, and a set of molecules $G_N$ generated from the model by sampling, we report the ratio of reproduced molecules $\frac{|G_N \cap T|}{|T|}$, and enrichment over random (EOR), which is defined as,

\begin{gather}
EOR = \frac{\frac{n}{|G_N|}}{\frac{m}{|R_M|}}
\end{gather}
where $n = |G_N \cap T|$ is the number of reproduced molecules from $T$ by sampling a set $G_N$ of $|G_N|=N$ molecules from the fine-tuned generative model, and $m = |R_M \cap T|$ is the number of reproduced molecules from $T$ by sampling a set $R_M$ of $|R_M|=M$ molecules from the generic, unbiased generative model trained only on the large dataset. Intuitively, EOR indicates how much better the fine-tuned models work when compared to the general model.

\section{Results and Discussion}

In this work, we address two points: First, we want to generate large sets of diverse molecules for virtual screening campaigns. Second, we want to generate smaller, focussed libraries enriched with possibly active molecules for a specific target.
For the first task, we can train a model on a large, general set of molecules to learn the \textsc{Smiles} grammar. Sampling from this model would generate sets of diverse, but unfocused molecules. To address the second task, and to obtain novel active drug molecules for a target of interest, we perform transfer learning: We select a small set of known actives for that target and we refit our pre-trained chemical language model with this small data-set. After each epoch, we sample from the model to generate novel actives. Furthermore, we investigate if the model actually benefits from transfer learning, by comparing it to a model trained from scratch on the small sets without pre-training.

\subsection{Training the recurrent network}
We employed a recurrent neural network with three stacked LSTM layers, each with 1024 dimensions, and each one followed by a dropout\cite{srivastava2014dropout} layer, with a dropout ratio of 0.2, to regularise the neural network. The model was trained until convergence, using a batch size of 128. The RNN was unrolled for 64 steps. It had $21.3 \times 10^6$ parameters.

\begin{table*}[htb]
\caption{Molecules sampled during training.}
\begin{center}
\begin{tabular}{rcr}
\toprule
Batch & Generated Example & valid\\
\midrule
0 & \texttt{Oc.BK5i\%ur+7oAFc7L3T=F8B5e=n)CS6RCTAR((OVCp1CApb)} & no\\
1000 & \texttt{OF=CCC2OCCCC)C2)C1CNC2CCCCCCCCCCCCCCCCCCCCCCC} & no\\
2000 & \texttt{O=C(N)C(=O)N(c1occc1OC)c2ccccc2OC} & yes\\
3000 & \texttt{O=C1C=2N(c3cc(ccc3OC2CCC1)CCCc4cn(c5c(Cl)cccc54)C)C} & yes\\
\bottomrule
\end{tabular}
\end{center}
\label{tab:gib}
\end{table*}%

During training, we sampled a few molecules from the model every 1000 mini-batches to inspect progress. Within a few 1000 steps, the model starts to output valid molecules (see Table \ref{tab:gib}).

\subsection{Generating Novel Molecules}
\label{sec:general}
To generate novel molecules, 50,000,000 \textsc{Smiles} symbols were sampled from the model symbol-by-symbol. This corresponded to 976,327 lines, from which 97.7\% were valid molecules after parsing with the CDK toolkit. 
Removing all molecules already seen during training yielded 864,880 structures. After filtering out duplicates, we obtained 847,955 novel molecules. 
A few generated molecules were randomly selected and depicted in Figure  \ref{fig:gentext}. The Supporting Information contains more structures. The created structures are not just formally valid, but are also mostly chemically reasonable. 

\begin{figure}[htbp]
\begin{center}
\includegraphics[width=0.9\linewidth]{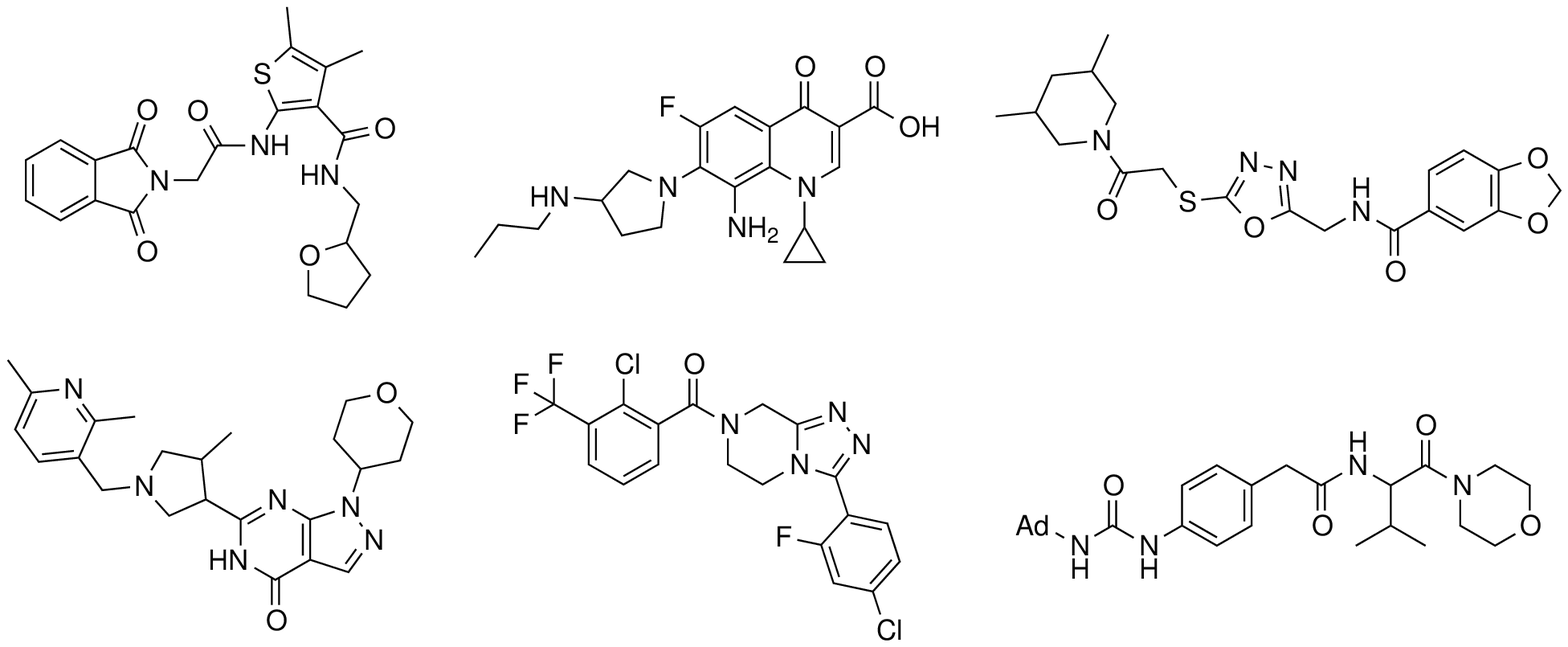}
\caption{A few randomly selected, generated molecules. Ad = Adamantyl}
\label{fig:gentext}
\end{center}
\end{figure}

In order to check if the \textit{de novo} compounds could be considered as valid starting points for a drug discovery program, we applied the internal AstraZeneca filters.\cite{cumming2013chemical} At AstraZeneca, this flagging system is used to determine if a compound is suitable to be part of the high-throughput screening collection (if flagged as ``core'' or ``backup'') or should be restricted for particular use (flagged as ``undesirable'' since it contains one or several unwanted substructures, e.g. undesired reactive functional groups). The filters were applied to the generated set of 848 k molecules and they flagged most of them, 640 k (75\%), are either core or backup. Since the same ratio (75\%) of core and backup compounds has been observed for the ChEMBL collection, we therefore conclude that the algorithm generates preponderantly valid screening molecules and faithfully reproduces the distribution of the training data.

\begin{figure}[htbp]
\begin{center}
\includegraphics[width=0.95\linewidth]{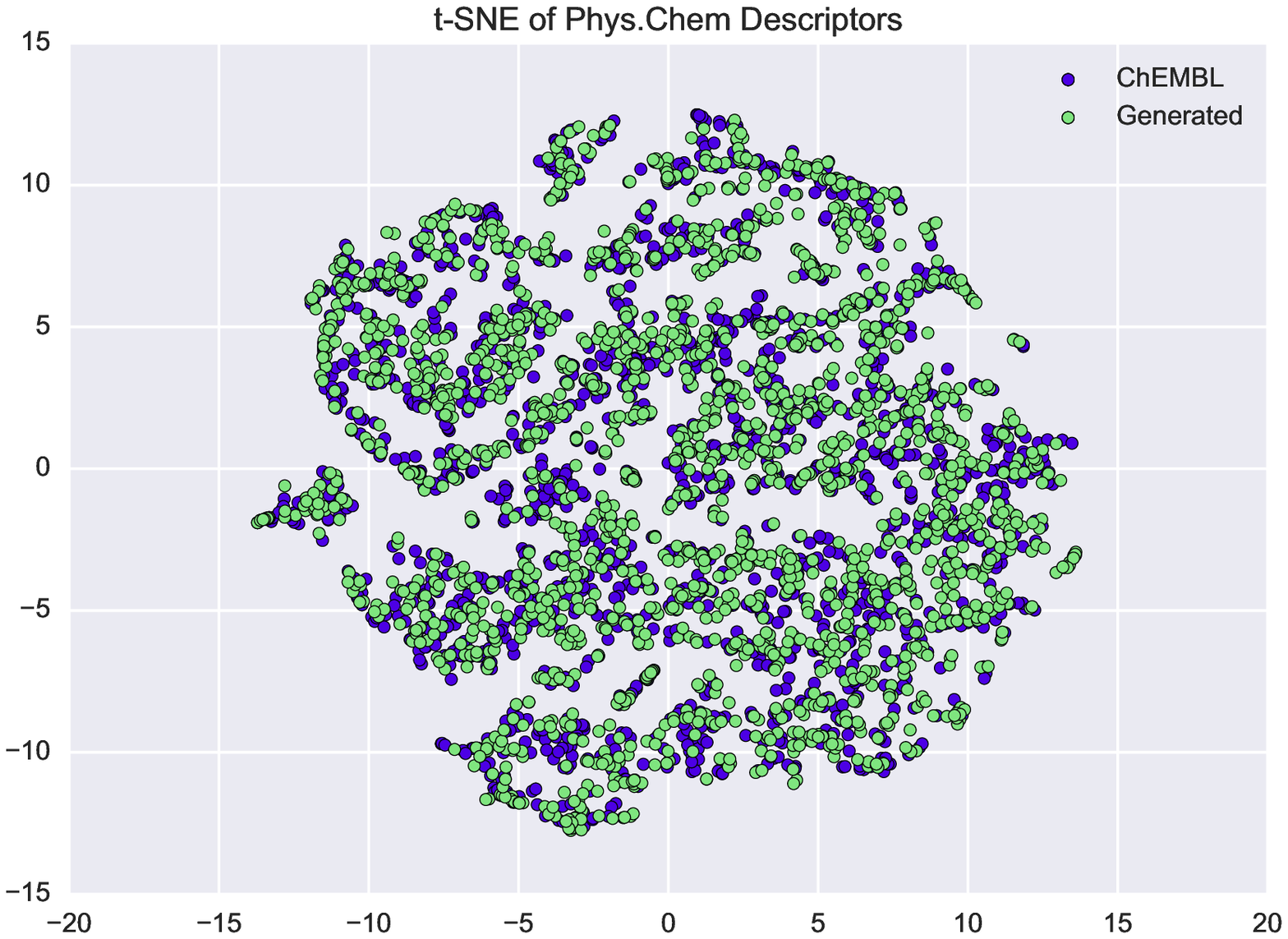}
\caption{t-SNE projection of 7 physicochemical descriptors of random molecules from ChEMBL (blue) and molecules generated with the neural network trained on ChEMBL (green), to two unitless dimensions. The distributions of both sets overlap significantly.}
\label{fig:tsne}
\end{center}
\end{figure}

To determine whether the properties of the generated molecules match the properties of the training data from ChEMBL, we followed the procedure of Kolb:\cite{chevillard2015scubidoo} We computed several molecular properties, namely molecular weight, BertzCT, the number of H-donors, H-acceptors, and rotatable bonds, logP and total polar surface area for randomly selected subsets from both sets with the RDKit\cite{rdkit} library version 2016.03.1. Then, we performed dimensionality reduction to 2D with t-SNE (t-Distributed Stochastic Neighbor Embedding, a technique analogous to PCA), which is shown in Figure \ref{fig:tsne}.\cite{maaten2008visualizing} Both sets overlap almost completely, which indicates that the generated molecules very well recreate the properties of the training molecules.

Furthermore, we analysed the Bemis-Murcko scaffolds of the training molecules and the sampled molecules.\cite{bemis1996properties} Bemis-Murcko scaffolds contain the ring systems of a molecule and the moieties that link these ring systems, while removing any side chains. They represent the scaffold, or ``core'' of a molecule, which series of drug molecules often have in common. The number of common scaffolds in both sets, divided by the union of all scaffolds in both sets (Jaccard index) is 0.12, which indicates that the language model does not just modify side chain substituents, but also introduces modifications at the molecular core.

\subsection{Generating Active Drug Molecules and Focused Libraries}
\subsubsection{Targeting the 5-HT$_{\mathbf{2A}}$ receptor}
To generate novel ligands for the 5-HT$_{\text{2A}}$ receptor, we first selected all molecules with \textit{p}IC$_{50}$ > 7 which were tested on 5-HT$_{\text{2A}}$ from ChEMBL (732 molecules), and then fine-tuned our pre-trained chemical language model on this set. After each epoch, we sampled 100,000 chars, canonicalised the molecules, and removed any sampled molecules that were already contained in the training set. Following this, we evaluated the generated molecules of each round of retraining with our 5-HT$_{\text{2A}}$ target prediction model (TPM). In Figure 4, the ratio of molecules predicted to be active by the TPM after each round of fine-tuning is shown. Before fine-tuning (corresponding to epoch 0), the model generates almost exclusively inactive molecules. Already after 4 epochs of fine-tuning the model produced a set in which 50\% of the molecules are predicted to be active.

\begin{figure}[htbp]
\begin{center}
\includegraphics[width=\linewidth]{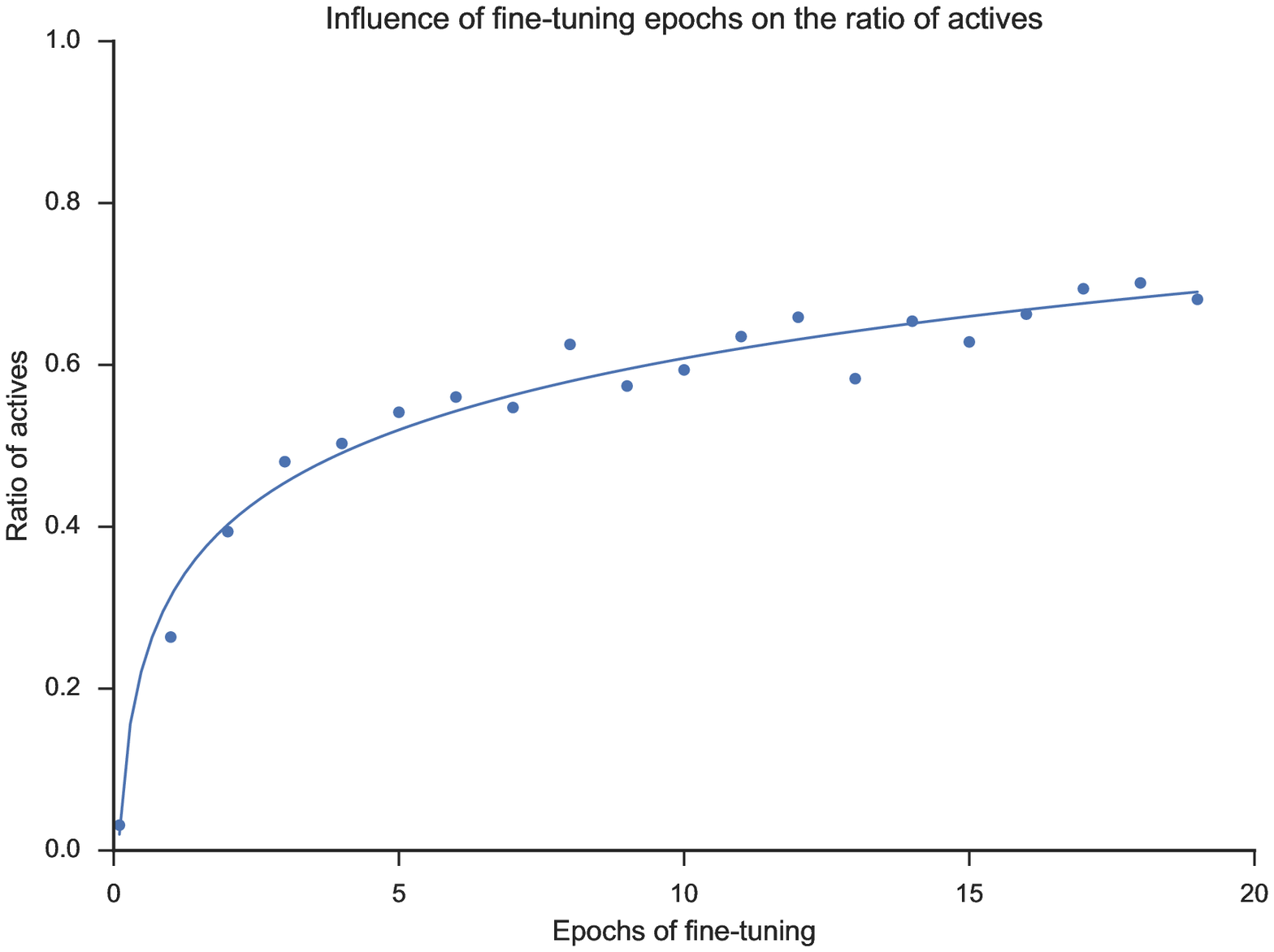}
\caption{Epochs of fine-tuning vs ratio of actives.}
\label{fig:5thpred}
\end{center}
\end{figure}

\begin{figure*}[htbp]
\begin{center}
\includegraphics[width=0.95\textwidth]{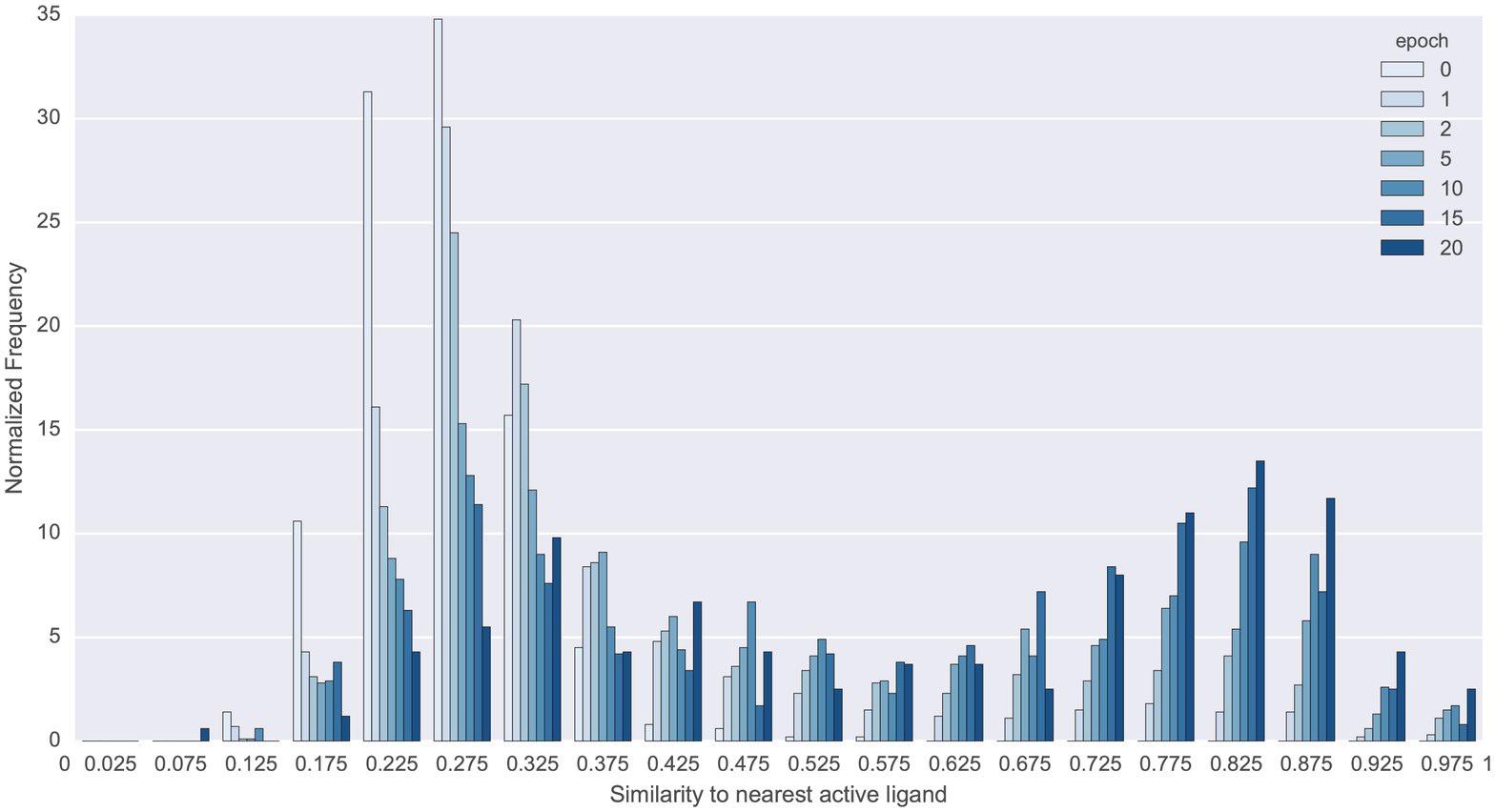}
\caption{Nearest-neighbour Tanimoto similarity distribution of the generated molecules for 5-HT$_{\text{2A}}$ after $n$ epochs of fine-tuning against the known actives. The generated molecules are distributed over the whole similarity range. Generated molecules with a medium similarity can be interesting for scaffold-hopping.\cite{stumpfe2011similarity}}
\label{fig:histo}
\end{center}
\end{figure*}

\paragraph{Diversity Analysis} 
In order to assess the novelty of the \textit{de novo} molecules generated with the fine-tuned model, a nearest neighbor similarity/diversity analysis has been conducted using a commonly used 2D fingerprint (ECFP4) based similarity method (Tanimoto index).\cite{chevillard2015scubidoo} Figure \ref{fig:histo} shows the distribution of the nearest neighbor Tanimoto index generated by comparing all the novel molecules and the training molecules before and after $n$ epochs of fine-tuning. For each bin, the white bars indicate the molecules generated from the unbiased, general model, while the darker bars indicate the molecules after several epochs of fine-tuning. Within the bins corresponding to lower similarity, the number of molecules decreases, while the bins of higher similarity get populated with increasing numbers of molecules. The plot thus shows that the model starts to output more and more similar molecules to the target-specific training set. Notably, after a few rounds of training not only highly similar molecules are produced, but also molecules covering the whole range of similarity, indicating that our method could not only deliver close analogs but new chemotypes or scaffold ideas to a drug discovery project.\cite{stumpfe2011similarity} To have the best of both worlds, that is diverse and focussed molecules, we therefore suggest to sample after each epoch of retraining and not just after the final epoch.

\subsubsection{Targeting Plasmodium falciparum (Malaria)}
\textit{Plasmodium falciparum} is a parasite that causes the most dangerous form of Malaria.\cite{williamson2016open} To probe our model on this important target, we used a more challenging validation strategy. We wanted to investigate whether the model could also propose the same molecules that medicinal chemists chose to evaluate in published studies. To test this, first, the known actives against Plasmodium falciparum with a \textit{p}IC$_{50}$ > 8 were selected from ChEMBL. Then, this set was split randomly into a training (1239 molecules) and a test set (1240 molecules). The chemical language model was then fine-tuned on the training set. 7500 molecules were sampled after each of the 20 epochs of refitting.  

\begin{table}[htb]
\caption{Reproducting known actives in the \textit{Plasmodium} test set. EOR: Enrichment over random.}
\begin{center}
\begin{tabular}{lrrrrrrr}
\toprule
\# & \textit{p}IC$_{50}$ & Train.& Test & Gen. mols. & Reprod. & EOR\\
\midrule
1 & > 8 & 1239 & 1240 & 128,256 &  28\%  & 66.9\\
2 & > 8 & 100  & 1240 & 93,721 &  7\%  & 19.0  \\
3 & > 9 &  100 & 1022 & 91,034  &  11\%  & 35.7\\
\bottomrule
\end{tabular}
\end{center}
\label{tab:malaria}
\end{table}%

\begin{figure}[htb]
\begin{center}
\includegraphics[width=0.95\linewidth]{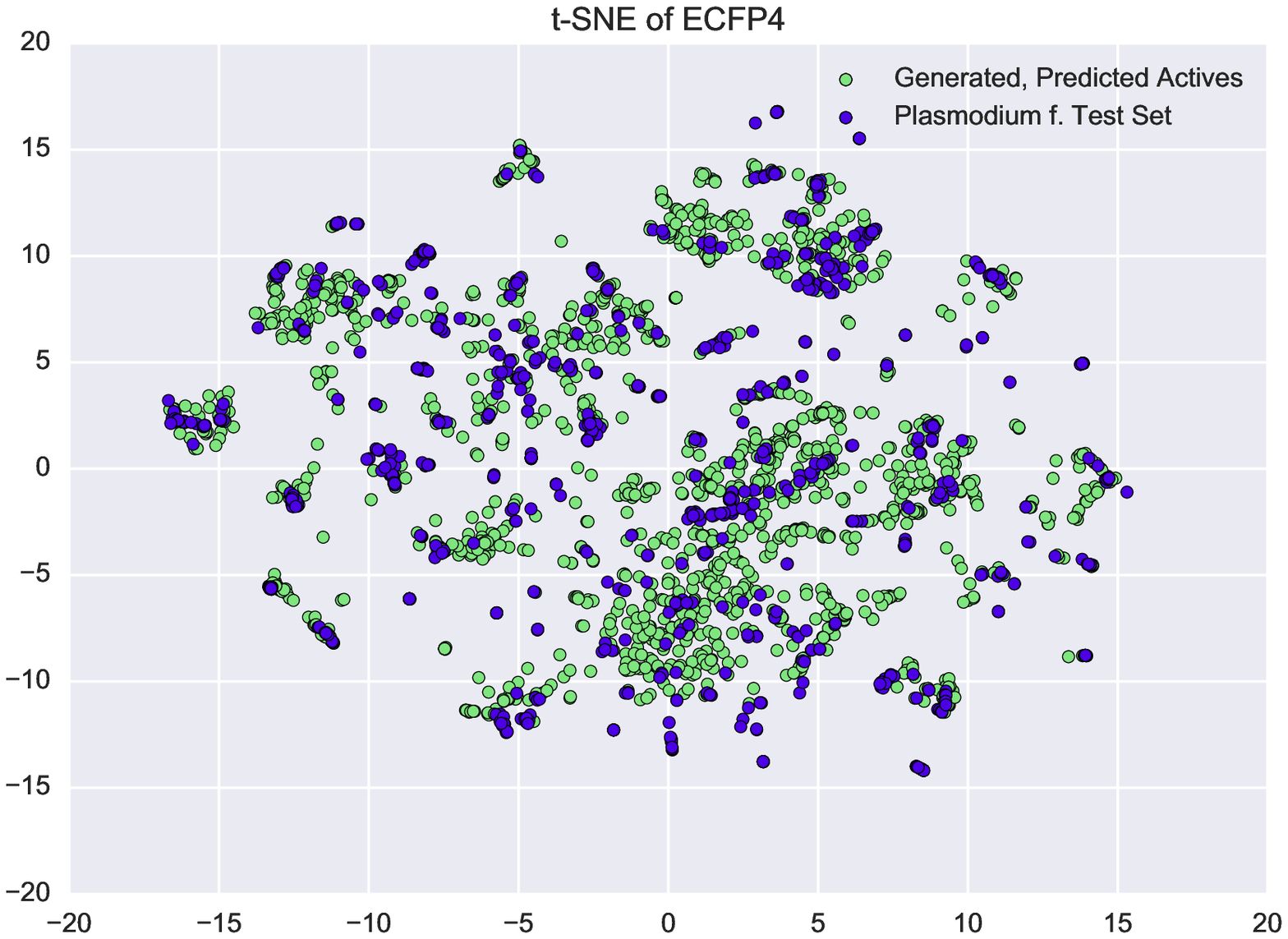}
\caption{t-SNE plot of the \textit{p}IC$_{50}$>9 test set (blue) and the \textit{de novo} molecules predicted to be active (green). The language model populates chemical space around the test molecules.}
\label{fig:tsnemalaria}
\end{center}
\end{figure}

This yielded 128,256 unique molecules. Interestingly, we found that our model was able to "redesign" 28\% of the unseen molecules of the test set. In comparison to molecules sampled from the unspecific, untuned model, an Enrichment over Random (EOR) of 66.9 is obtained. With a smaller training set of 100 molecules, the model can still reproduce 7\% of the test set, with an EOR of 19.0. 
To test the reliance on \textit{p}IC$_{50}$ we chose to use another cut-off of \textit{p}IC$_{50}$ > 9, and took 100 molecules in the training set and 1022 in the test set. 11\% of the test set could be recreated, with an EOR of 35.7. To visually explore how the model populates chemical space, Figure \ref{fig:tsnemalaria} shows a t-SNE plot of the ECFP4 fingerprints of the test molecules and 2000 generated molecules that were predicted to be active by the target prediction model for \textit{Plasmodium falciparum}. It indicates that the model has generated many similar molecules around the test examples.

\subsubsection{Targeting Staphylococcus aureus (Golden Staph)}

\begin{figure}[htbp]
\begin{center}
\includegraphics[width=\linewidth]{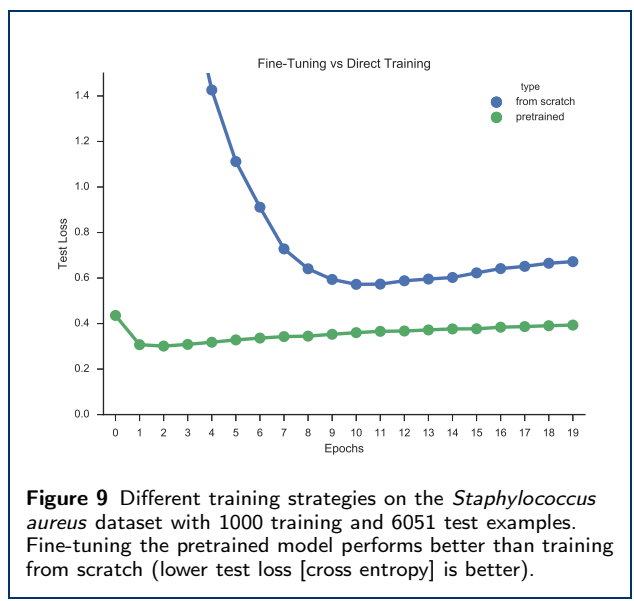}
\caption{Different training strategies on the \textit{Staphylococcus aureus} dataset with 1000 training and 6051 test examples. Fine-tuning the pretrained model performs better than training from scratch (lower test loss [cross entropy] is better).}
\label{fig:pretraining}
\end{center}
\end{figure}

\begin{table*}[htb]
\caption{Reproducting known actives in the \textit{Staphylococcus} test set. EOR: Enrichment over random.}
\begin{center}
\begin{tabular}{lrrrrrr}
\toprule
Entry  & \textit{p}MIC & Train.& Test & Gen. mols. & Reprod. & EOR\\
\midrule
1  & > 3 & 1000 & 6051 & 51,052  &  14\%  & 155.9 \\
2  & > 3 &   50 & 7001 & 70,891  &  2.5\% & 21.6  \\
3$^a$ & > 3 &50 & 7001& 85,755  &  1.8\% &6.3    \\
4$^b$ & > 3 &50 & 7001&    285  &  0\% & ---      \\
5$^c$ & > 3 &0 & 7001&  60,988  &  6\% &  59.6   \\
\bottomrule
\end{tabular}
\\
$^a$Fine-tuning learning rate = $10^{-4}$. $^b$No Pretraining. $^c$8 Generate-Test cycles.
\end{center}
\label{tab:aureus}
\end{table*}%
To evaluate a different target, we furthermore conducted a series of experiments to reproduce known active molecules against \textit{Staphylococcus aureus}. Here, we used actives with a \textit{p}MIC > 3. MIC is the Mean Inhibitory Concentration, the lowest concentration of a compound that prevents visible growth of a microorganism. As above, the actives were split into a training and a test set. However, here, the availability of the data allows larger test sets to be used. After fine-tuning on the training set of 1000 molecules (Table \ref{tab:aureus}, Entry 1), our model could retrieve 14\% of the 6051 test molecules.  
When scaling down to a smaller training set of 50 molecules (the model gets trained on less than 1\% of the data!), it can still reproduce 2.5\% of the test set, and performs 21.6 times better than the unbiased model (Table \ref{tab:aureus}, Entry 2). Using a lower learning rate (0.0001, Entry 3) for fine-tuning, which is often done in transfer learning, does not work as well as the standard learning rate (0.001, Entry 2). We additionally examined whether the model benefits from transfer learning. When trained from scratch, the model performs much worse than the pretrained and subsequently fine-tuned model (see Figure \ref{fig:pretraining} and Table \ref{tab:aureus}, Entry 4). Pretraining on the large dataset is thus crucial to achieve good performance against \textit{Staphylococcus aureus}.

\subsection{Simulating Design-Synthesis-Test Cycles}
\begin{figure}[htbp]
\begin{center}
\includegraphics[width=0.95\linewidth]{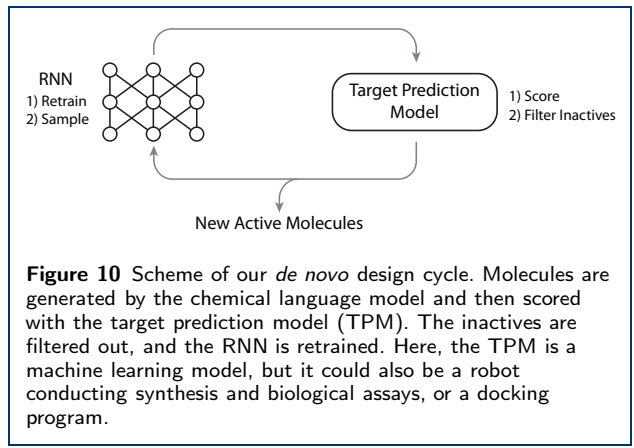}
\caption{Scheme of our \textit{de novo} design cycle. Molecules are generated by the chemical language model and then scored with the target prediction model (TPM). The inactives are filtered out, and the RNN is retrained. Here, the TPM is a machine learning model, but it could also be a robot conducting synthesis and biological assays, or a docking program.}
\label{fig:designcycle}
\end{center}
\end{figure}

The experiments we conducted so far are applicable if one already knows several actives. However, in drug discovery, one often does not have such a set to start with. Therefore, high throughput screenings are conducted to identify a few hits, which serve as a starting point for the typical cyclical drug discovery process: Molecules get designed, synthesised, and then tested in assays. Then, the best molecules are selected, and based on the gained knowledge new molecules are designed, which closes the cycle. Therefore, as a final challenge for our model, we simulated this cycle by iterating molecule generation ("synthesis"), selection of the best molecules with the machine learning-based target prediction ("virtual assay") and retraining the language model with the best molecules ("design") with \textit{Staphylococcus aureus} as the target. We thus do not use a set of known actives to start the structure generation procedure (see Figure \ref{fig:designcycle}).

We started with 100,000 sampled molecules from the unbiased chemical language model. Then, using our target prediction model, we extracted the molecules classified as actives. After that, the RNN was fine-tuned for 5 epochs on the actives, sampling $\approx$10,000 molecules after each epoch. The resulting molecules were filtered with the target prediction model, and the new actives appended to the actives from the previous round, closing the loop.

Already after 8 iterations, the model reproduced 416 of the 7001 test molecules from the previous task, which is 6\% (Table \ref{tab:aureus}, Entry 5), and exhibits and EOR of 59.6. This EOR is higher than if the model is retrained directly on a set of 50 actives (Entry 2). Additionally, we obtained 60,988 unique molecules that the target prediction model classified as active. This demonstrates that in combination with a target prediction or scoring model, our model can also perform the complete \textit{de novo}-design cycle.

\subsection{Why does the model work?}
Our results presented in Section \ref{sec:general} show that the general model trained on a large molecule set has learned the \textsc{Smiles} rules and can output valid, drug-like molecules, which resemble the training data.
However, sampling from this model does not help much if we want to generate actives for a specific target: We would have to generate very large sets to find actives for that target among the diverse range of molecules the model creates, which is indicated by the high EOR scores in our experiments.

\begin{figure}[htbp]
\begin{center}
\includegraphics[width=\linewidth]{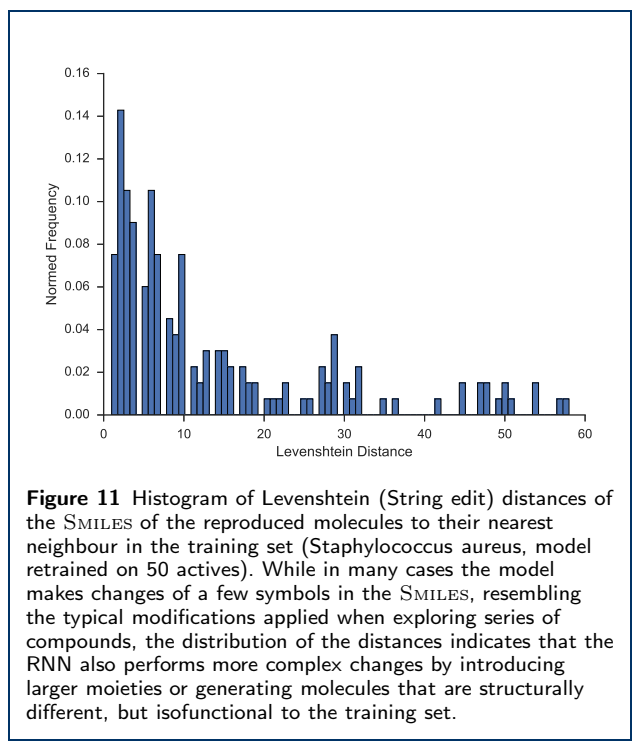}
\caption{Histogram of Levenshtein (String edit) distances of the \textsc{Smiles} of the reproduced molecules to their nearest neighbour in the training set (Staphylococcus aureus, model retrained on 50 actives). While in many cases the model makes changes of a few symbols in the \textsc{Smiles}, resembling the typical modifications applied when exploring series of compounds, the distribution of the distances indicates that the RNN also performs more complex changes by introducing larger moieties or generating molecules that are structurally different, but isofunctional to the training set.}
\label{fig:levenshtein}
\end{center}
\end{figure}

\begin{figure}[htbp]
\begin{center}
\includegraphics[width=\linewidth]{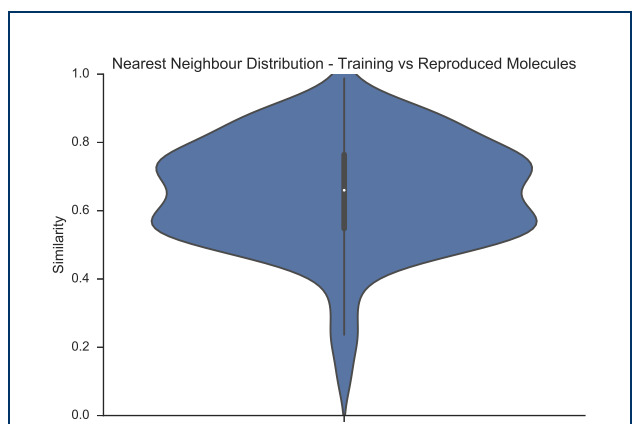}
\caption{Violin plot of the nearest-neighbour ECFP4-Tanimoto similarity distribution of the 50 training molecules against the rediscovered molecules in Table \ref{tab:aureus}, Entry 2. The distribution suggests the model has learned to make typical small functional group replacements, but can also reproduce molecules which are not too similar to the training data.}
\label{fig:repr-train-nn}
\end{center}
\end{figure}

When fine-tuned to a set of actives, the probability distribution over the molecules captured by our model is shifted towards molecules active towards our target. To study this, we compare the Levenshtein (String edit) distance of the generated \textsc{Smiles} to their nearest neighbours in the training set in Figure \ref{fig:levenshtein}. The Levenshtein distance of e.g. benzene \texttt{c1ccccc1} and pyridine \texttt{c1ccncc1} would be 1. Figure \ref{fig:levenshtein} shows that while the model often seems to have made small replacements in the underlying \textsc{Smiles}, in many cases it also made more complex modifications or even generated completely different \textsc{Smiles}. This is supported also by the distribution of the nearest neighbour fingerprint similarities of training and rediscovered molecules (ECFP4, Tanimoto, Figure \ref{fig:repr-train-nn}). Many rediscovered molecules are in the medium similarity regime.

Because we perform transfer learning, during fine-tuning, the model does not "forget" what it has learned. A plausible explanation why the model works is therefore that it can transfer the modifications that are regularly applied when series of molecules are studied, to the molecules it has seen during fine-tuning.

\section{Conclusion}
In this work, we have shown that recurrent neural networks based on the Long Short Term Memory (LSTM) can be applied to learn a statistical chemical language model. The model can generate large sets of novel molecules with similar physico-chemical properties to the training molecules. This can be used to generate libraries for virtual screening. Furthermore, we demonstrated that the model performs transfer learning when fine-tuned to smaller sets of molecules active towards a specific biological target, which enables the creation of novel molecules with the desired activity. By iterating cycles of structure generation with the language model, scoring with a target prediction model (TPM) and retraining of the model with increasingly larger sets of highly scored molecules, we showed that we do not even need a set of active known active molecules to start our procedure with, as the TPM could also be a docking program, or a robot conducting synthesis\cite{ley2015organic} and biological testing.

We see three main advantages of our method. First, it is conceptually orthogonal to established molecule generation approaches, as it learns a generative model for molecular structures. Second, our method is very simple to setup, train and to use, and can be adapted to different datasets without any modifications to the model architecture, and does not depend on hand-encoded expert knowledge. Furthermore, it merges structure generation and optimisation in one model.
A weakness of our model is interpretability. In contrast, existing de-novo design methods settled on virtual reactions to generate molecules, which has advantages as it minimises the chance of obtaining "overfit", weird molecules, and increases the chances to find synthesizable compounds.\cite{hartenfeller2011enabling,schneider2016novo} 

To extend our work, it is just a small step to cast molecule generation as a reinforcement learning problem, where the pre-trained LSTM generator could be seen as a policy, which can be encouraged to create better molecules with a reward signal obtained from a target prediction model.\cite{sutton1998reinforcement} In addition, different approaches for target prediction, for example docking, could be evaluated.\cite{kitchen2004docking,hartenfeller2011enabling}

Deep Learning is not a panacea, and we join Gawehn et al. in expressing ``some healthy skepticism'' regarding its application in drug discovery.\cite{gawehn2016deep} Generating molecules that are almost right is not enough, because in Chemistry, a miss is as good as a mile, and drug discovery is a ``needle in the haystack'' problem -- in which also the needle looks like hay. 
Nevertheless, given that we have shown in this work that our model can rediscover those needles, and other recent developments,\cite{gawehn2016deep,schmidhuber2015deep,altae2016low,graves2016hybrid} we believe that deep neural networks can be complimentary to established approaches in drug discovery. The complexity of the problem certainly warrants the investigation of novel approaches. Eventually, success in the wet lab will determine if the new wave\cite{zupan1999neural} of neural networks will prevail.

%%%%%%%%%%%%%%%%%%%%%%%%%%%%%%%%%%%%%%%%%%%%%%
%%                                          %%
%% Backmatter begins here                   %%
%%                                          %%
%%%%%%%%%%%%%%%%%%%%%%%%%%%%%%%%%%%%%%%%%%%%%%

\begin{backmatter}

\section*{Competing interests}
  The authors declare that they have no competing interests.

%\section*{Author's contributions}%
%

\section*{Acknowledgements}
The project was conducted during a research stay of M.S. at AstraZeneca R\&D Gothenburg. We thank H. Chen and O. Engkvist for valuable discussions and feedback on the manuscript, and G. Klambauer for helpful suggestions.
%%  After submission of the .TEX file,                     %%
%%  you will be prompted to submit your .BBL file.         %%

% if your bibliography is in bibtex format, use those commands:
%\bibliographystyle{bmc-mathphys} % Style BST file (bmc-mathphys, vancouver, 
%spbasic).
\bibliographystyle{achemso}
\bibliography{lstm}      % Bibliography file (usually '*.bib' )

\section*{Additional Files}
%  \subsection*{Additional file 1 --- Sample additional file title}
%%    Additional file descriptions text (including details of how to
 %   view the file, if it is in a non-standard format or the file extension).  This might
 %   refer to a multi-page table or a figure.

  \subsection*{Generated Molecules}
In the following, a few randomly selected molecules produced with the general model trained on ChEMBL are shown. 
\begin{figure*}[htbp]
\begin{center}
\includegraphics[width=0.95\textwidth]{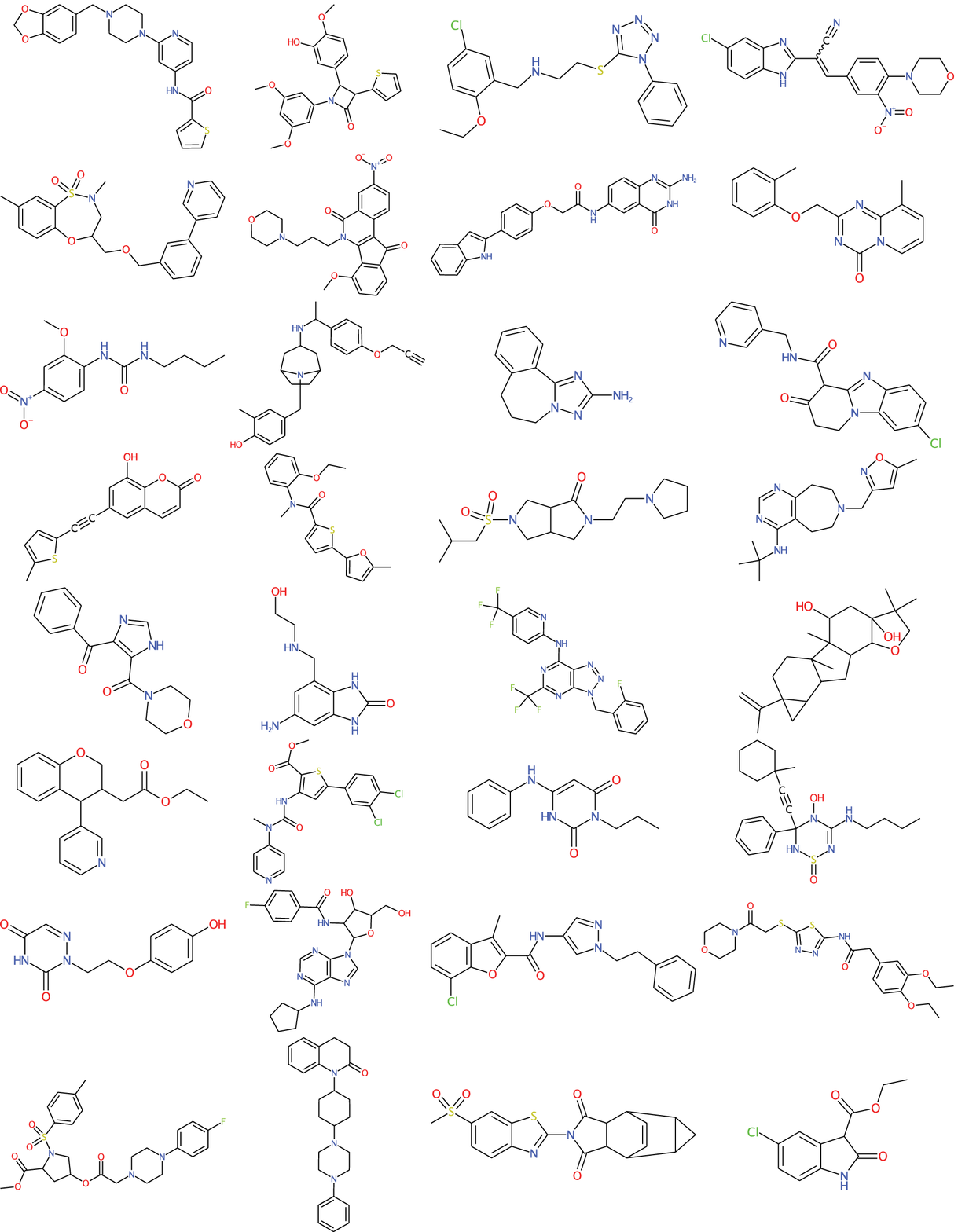}
\end{center}
\end{figure*}

\begin{figure*}[htbp]
\begin{center}
\includegraphics[width=0.95\textwidth]{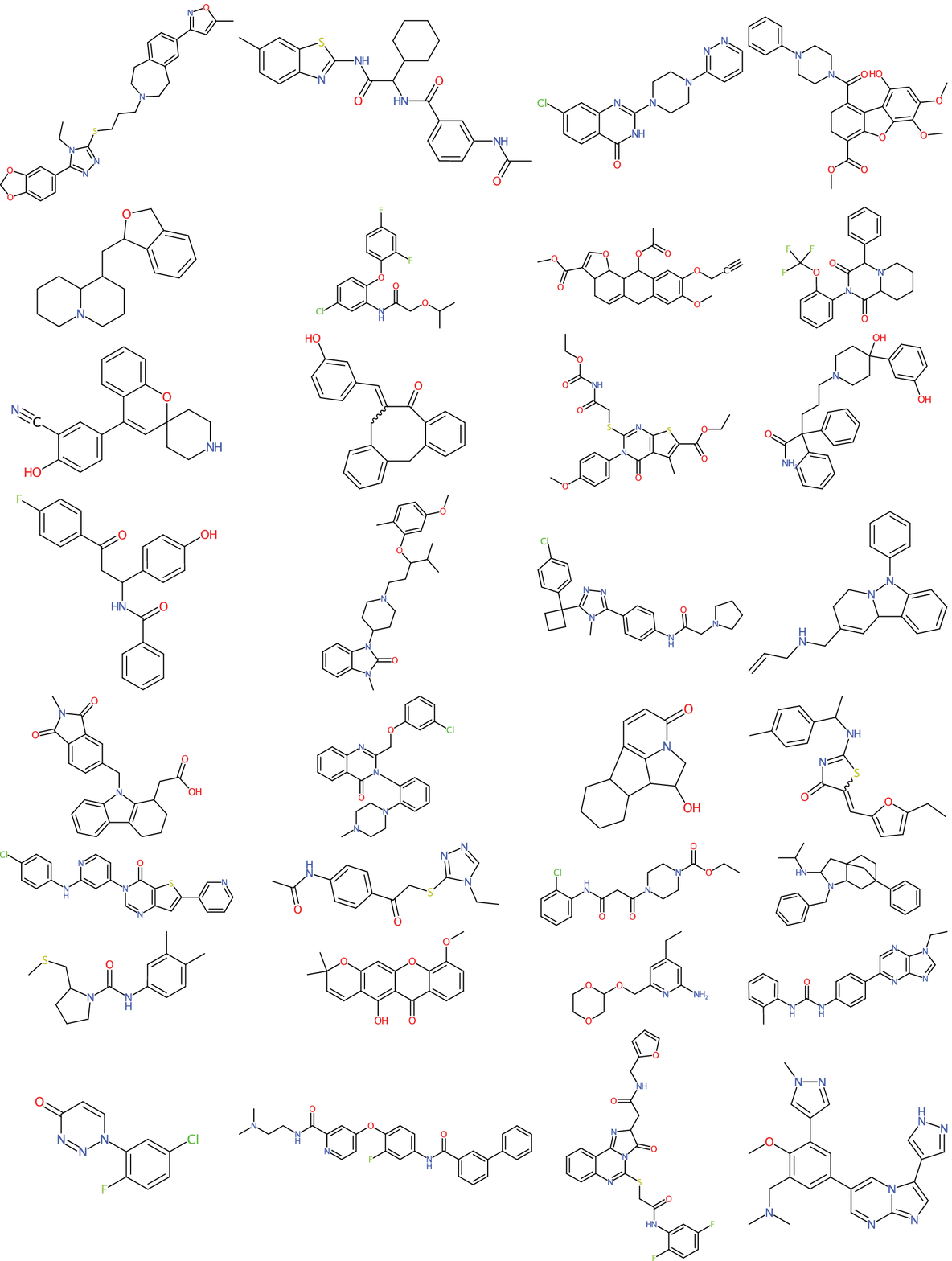}
\end{center}
\end{figure*}

\begin{figure*}[htbp]
\begin{center}
\includegraphics[width=0.95\textwidth]{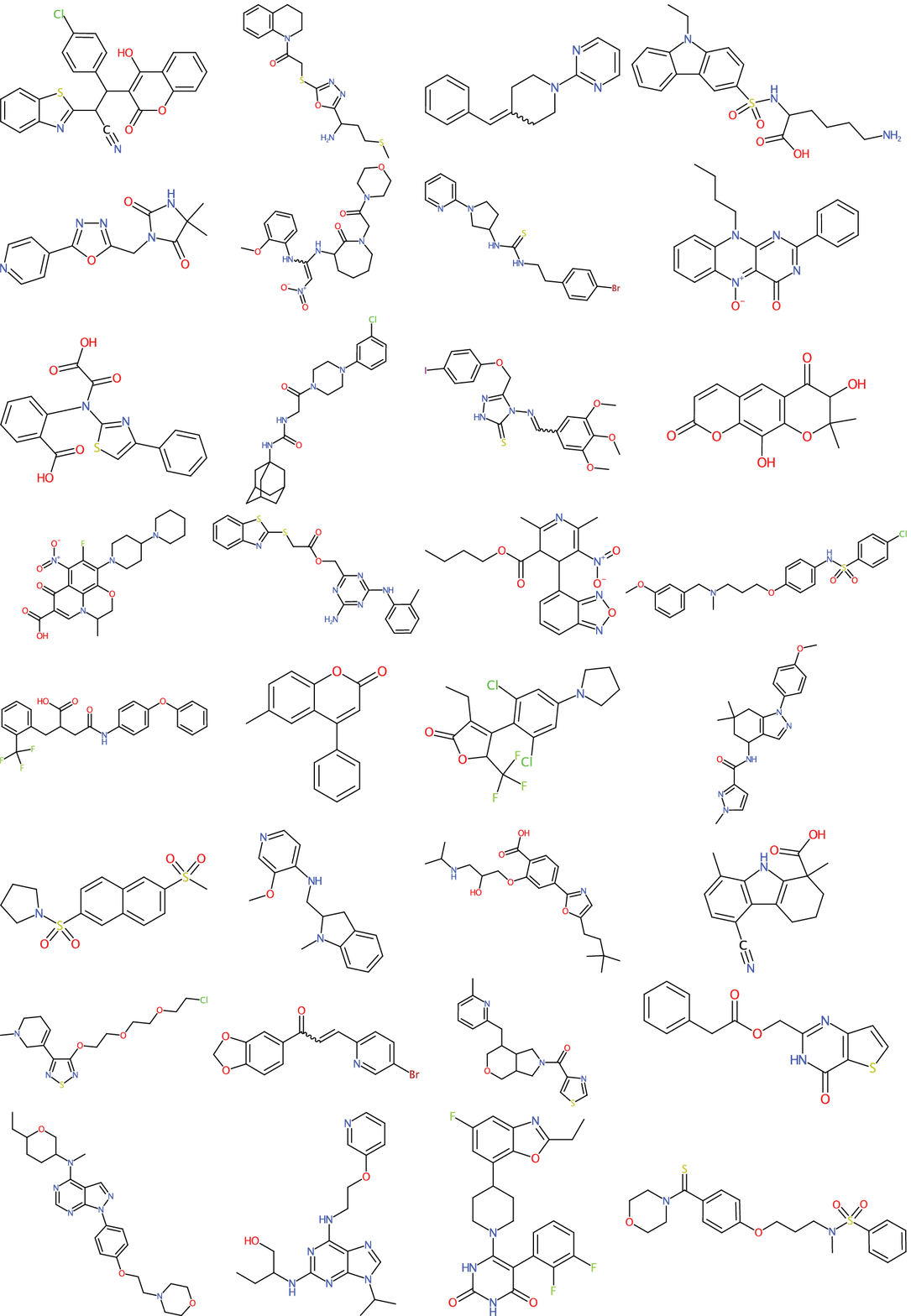}
\end{center}
\end{figure*}

\begin{figure*}[htbp]
\begin{center}
\includegraphics[width=0.95\textwidth]{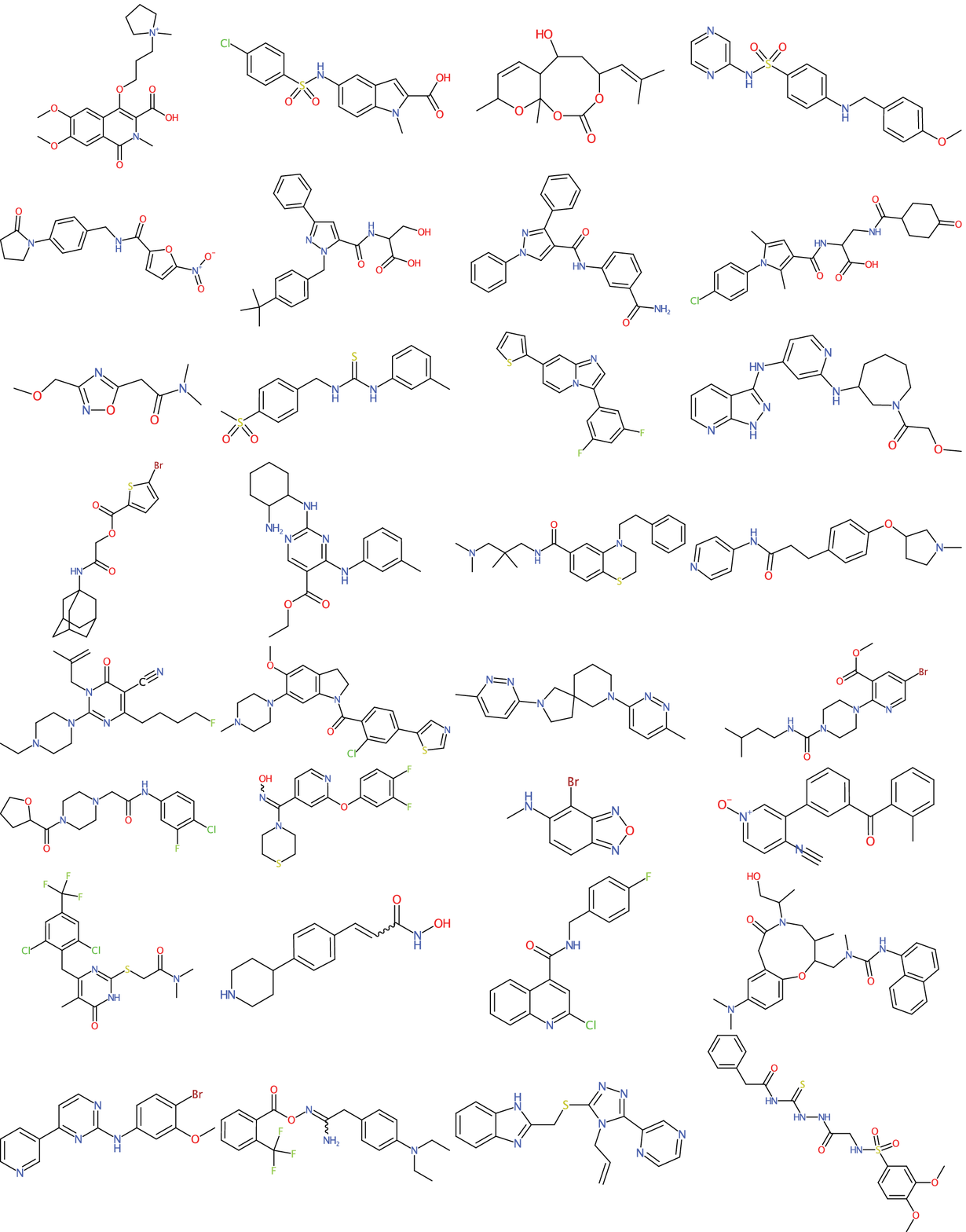}
\end{center}
\end{figure*}

\begin{figure*}[htbp]
\begin{center}
\includegraphics[width=0.95\textwidth]{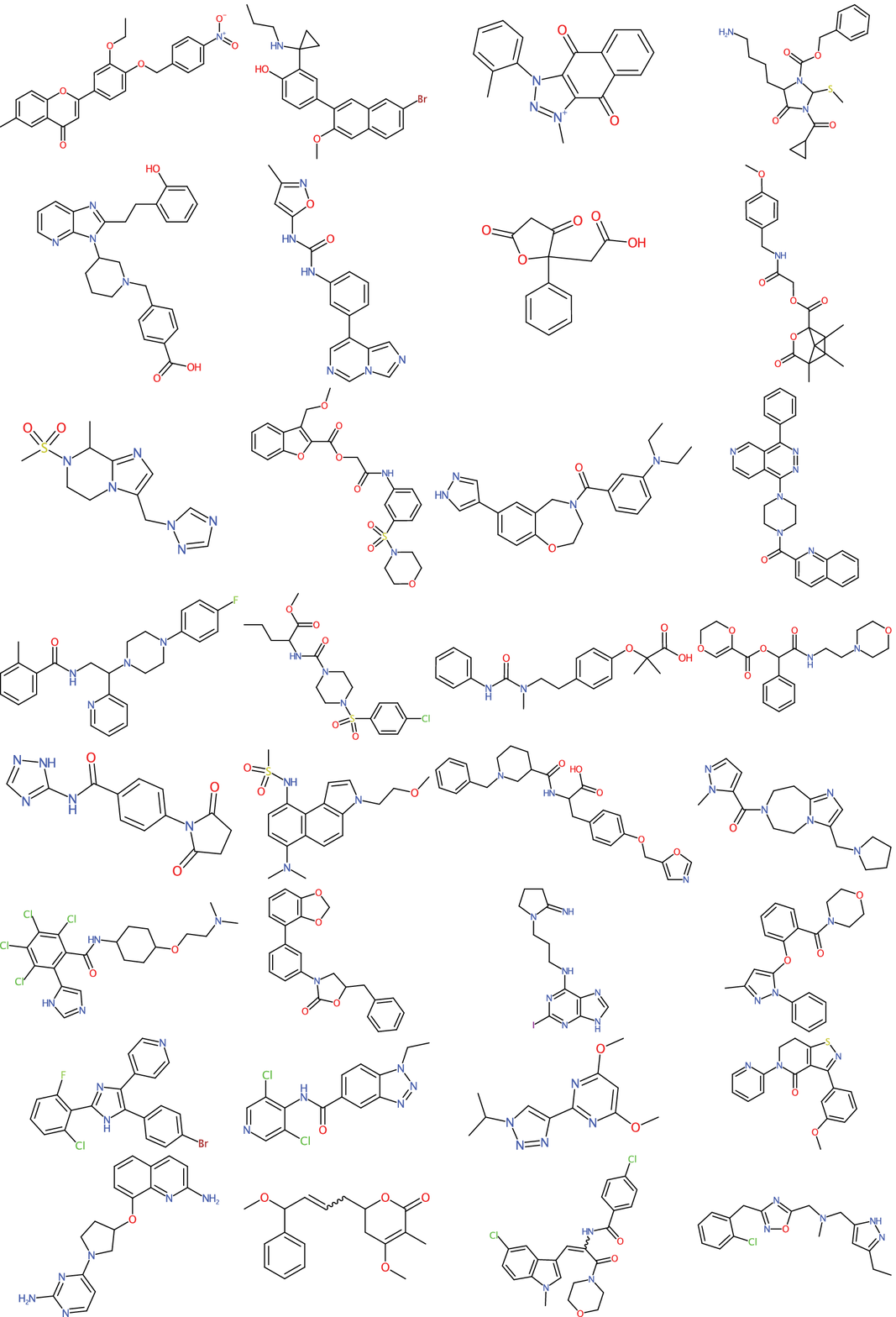}
\end{center}
\end{figure*}

\end{backmatter}
\end{document}